\documentclass[sigconf]{acmart}

\usepackage{balance}

\def\BibTeX{{\rm B\kern-.05em{\sc i\kern-.025em b}\kern-.08emT\kern-.1667em\lower.7ex\hbox{E}\kern-.125emX}}
    
\setcopyright{acmcopyright}
\copyrightyear{2019} 
\acmYear{2019} 
\acmConference[MM '19]{Proceedings of the 27th ACM International Conference on Multimedia}{October 21--25, 2019}{Nice, France}
\acmBooktitle{Proceedings of the 27th ACM International Conference on Multimedia (MM '19), October 21--25, 2019, Nice, France}
\acmPrice{15.00}
\acmDOI{10.1145/3343031.3350996}
\acmISBN{978-1-4503-6889-6/19/10}

\acmSubmissionID{fp675}

\usepackage{algorithm}  
\usepackage{algpseudocode}  
\usepackage{amsmath}
\usepackage{multirow}
\begin{document}


\fancyhead{}

\title{Unpaired Cross-lingual Image Caption Generation with Self-Supervised Rewards}

\author{Yuqing Song}
\affiliation{%
 \institution{Renmin University of China}
}
\email{syuqing@ruc.edu.cn}

\author{Shizhe Chen}
\affiliation{%
 \institution{Renmin University of China}
}
\email{cszhe1@ruc.edu.cn}

\author{Yida Zhao}
\affiliation{%
 \institution{Renmin University of China}
}
\email{zyiday@ruc.edu.cn}

\author{Qin Jin}
\authornote{Corresponding Author}
\affiliation{%
 \institution{Renmin University of China}
}
\email{qjin@ruc.edu.cn}

\renewcommand{\shortauthors}{Song, et al.}

\begin{abstract}
Generating image descriptions in different languages is essential to satisfy users worldwide.
However, it is prohibitively expensive to collect large-scale paired image-caption dataset for every target language which is critical for training descent image captioning models.
Previous works tackle the unpaired cross-lingual image captioning problem through a pivot language, which is with the help of paired image-caption data in the pivot language and pivot-to-target machine translation models.
However, such language-pivoted approach suffers from inaccuracy brought by the pivot-to-target translation, including disfluency and visual irrelevancy errors.
In this paper, we propose to generate cross-lingual image captions with self-supervised rewards in the reinforcement learning framework to alleviate these two types of errors.
We employ self-supervision from mono-lingual corpus in the target language to provide fluency reward, and propose a multi-level visual semantic matching model to provide both sentence-level and concept-level visual relevancy rewards.
We conduct extensive experiments for unpaired cross-lingual image captioning in both English and Chinese respectively on two widely used image caption corpora.
The proposed approach achieves significant performance improvement over state-of-the-art methods. 

\end{abstract}


\maketitle

\section{Introduction}
Generating natural language sentences to describe the image content, a.k.a image captioning, has received more and more attention in recent years.
It could help visually impaired people to better understand the real world, and make it easier to index and retrieve massive images on the web.
Thanks to the rapid development of computer vision and natural language generation, remarkable progress has been made in automatic image captioning. 
However, most of previous works have mainly focused on generating English captions for images. 
As we know, there are more than 6.6 billion non-native English speakers in the world, and the benefits of image captioning technology should also be brought to these users.
Therefore, it is necessary to generate captions in different languages, which is also called the cross-lingual image captioning task.

\begin{figure}
  \centering
  \includegraphics[width=\linewidth]{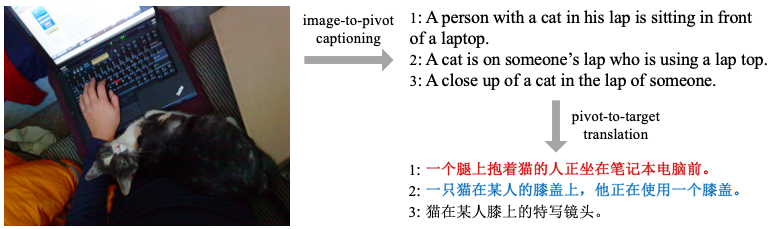}
  \caption{Illustration of cross-lingual Chinese image captioning with English language as pivot. The translated Chinese caption in red suffers from disfluency errors while sentence in blue contains visual irrelevancy errors.}
  \Description{Case of cross-lingual image captioning}
  \label{example}
\end{figure}

Since image captioning models are generally data-hungry, the main challenge for cross-lingual image captioning is the lack of large-scale image caption dataset in the target language.
It is also prohibitively expensive to collect dataset for each language.
Fortunately, great efforts have already been made in collecting large-scale image-caption datasets in English, as well as machine translation datasets from English to other languages.
Therefore, for cross-lingual image captioning, a natural way to avoid the demand of paired image-caption data in the target language is to employ another language, such as English, as the pivot to bridge the image and the target language \cite{gu2018pivot}, so that the image caption is first generated by an image-to-pivot captioning model, and then translated into the target language by a pivot-to-target machine translation (MT) model.
Figure~\ref{example} illustrates the idea of utilizing English as pivot for cross-lingual Chinese image captioning.

The major limitation of such language-pivoted approach is that translation errors brought by the pivot-to-target MT model cannot be corrected which seriously affects the quality of generated captions, especially when the MT model is trained in a different domain from image captions.
In order to alleviate the domain mismatch, Gu \emph{et al.} \cite{gu2018pivot} propose to share parameters between the image-to-pivot captioning model and the pivot-to-target MT model, and jointly train these two models, which can enforce the MT model to adapt styles towards image captions.
However, this approach is hard to generalize and computationally expensive to employ state-of-the-art MT models for pivot-to-target translation.
Lan \emph{et al.} \cite{lan2017fluencyguided} instead directly take advantage of the state-of-the-art translator \footnote{Baidu online translation: http://api.fanyi.baidu.com} to generate pseudo image-target caption pairs to train the image-to-target captioning model.
They propose to re-weight the translated captions by the language fluency.
However, in addition to the disfluent sentences, the imperfect translations may also contain fluent but visually irrelevant sentences as shown in Figure~\ref{example}, which also greatly affect the accuracy of cross-lingual caption generation.

In this paper, we propose a self-supervised rewarding model (SSR) to deal with both disfluency and visual irrelevancy errors in language-pivoted unpaired image captioning.
Our model is based on the reinforcement learning framework, which utilizes two types of rewards learned from self-supervisions to encourage the caption generator to correct above errors.
Specifically, to improve the caption fluency, we propose a fluency reward based on a target language model, which is trained with self-supervision loss on mono-lingual sentences in the target language.
In order to improve the visual relevancy, we propose a multi-level visual semantic matching model (ML-VSE) to provide relevancy rewards, which employs self-supervised pseudo image-target caption pairs from the pivot-to-target translation model for training.
The ML-VSE model contains both sentence-level and concept-level visual semantic matching between images and captions, which provides coarse- and fine-grained rewards respectively.
Extensive experiments on two widely used image caption datasets show that our model significantly outperforms prior works on all the caption performance metrics.

The main contributions of this work are summarized as follows:
\begin{itemize}
    \item We propose to employ the reinforcement learning framework to deal with errors in language-pivoted approaches for unpaired cross-lingual image captioning.
    \item Introspective self-supervisions with respect to the fluency and visual relevancy of generated captions are designed as the rewards to improve the quality of cross-lingual captions.
    \item Extensive experiments for both unpaired English image captioning and Chinese image captioning demonstrate that our proposed approach achieves significant improvement over previous methods on both objective caption metrics and human evaluation. 
\end{itemize}

\section{Related works}
\subsection{Image Caption Generation}
Image caption generation is a challenging task which connects computer vision and natural language processing. 
With the rapid development in deep learning, great breakthroughs have been made in image captioning \cite{fang2015captionback, Jia2015lstmforcaption, vinyals2015showtell, Hitschler2016multimodalpivot, zhang2017imagecaption, liu2018imagecaption}. 
Vinyals \emph{et al.} \cite{vinyals2015showtell} first propose an end-to-end image captioning model based on the encoder-decoder framework \cite{cho2014encoderdecoder}. 
A convolutional neural network (CNN) \cite{kim2014cnn} is used to encode the image into a fix-dimensional feature vector and a recurrent neural network (RNN) \cite{Sepp1997lstm} is used  as the decoder to generate captions based on the encoder output. 
The model is jointly optimized by maximizing the log probability of groundtruth descriptions.

Later, many improvements based on such encoder-decoder framework are proposed. 
Xu \emph{et al.} \cite{xu2015showatttell} propose the spatial attention mechanism for image captioning, which divides the image into grids, and teaches the model to attend to the corresponding grid at each decoding step. Anderson \emph{et al.} \cite{Anderson2017updown} replace the grids with detected objects in a bottom-up attention to enhance the previous top-down attention method. 
You \emph{et al.} \cite{you2016semanticatt} propose semantic attention which pre-defines a list of visual concepts to be attended to in the decoding step.
Gu \emph{et al.} \cite{gu2017cnndecoder} propose to explore both long-term and temporal information in captions with a CNN-based image captioning model.
Recently, Biten \emph{et al.} \cite{biten2019goodnews} propose to integrate contextual information into the captioning pipeline to deal with the out-of-vocabulary named entity generation.

Besides model structures, the training target also plays an important role in image captioning.
The model trained by traditional maximum likelihood target suffers from exposure bias and evaluation mismatch.
The exposure bias is caused by the training setting called ``Teacher-Forcing'' \cite{bengio2015teacherforce}, where the model has never been exposed to its own predictions in the training progress, which results in the error accumulation at test time. 
The evaluation mismatch exists because cross entropy is used as the training loss, but metrics such as BLEU, CIDEr and METEOR are instead used for caption performance evaluation. 
Therefore, reinforcement learning approaches are proposed to address these two problems.  
Rennie \emph{et al.} \cite{rennie2017sc} propose a new training method based on reinforcement learning with a baseline reward called ``Self-Critical''. They provide ``reward'' for captions sampled from model distribution, and the reward is directly evaluated by CIDEr. 
In order to enhance the stability of training, they use the reward of captions generated at test time as a baseline reward.
Works in \cite{luo2018discrim, liu2018showdiscrim} propose to train the captioning model by providing rewards of discriminability to improve the diversity of generated captions.
Our training strategy is similar to Rennie \emph{et al.} \cite{rennie2017sc} except that we use self-supervision with respect to fluency and relevancy as rewards for model learning. 

\vspace{-3pt}
\subsection{Cross-lingual Image Captioning}
Cross-lingual image captioning is a more challenging captioning task which has not been well investigated yet, since most previous works have mainly focused on generating English captions.
Tsutsui \emph{et al.} \cite{Tsutsui2017multilingual} propose to generate image captions in Japanese by collecting a large-scale parallel image-caption dataset in Japanese. 
However, it may not be feasible for many languages due to the expensive cost of dataset collection.
Feng \emph{et al.} \cite{feng2018unsupervised} propose an unsupervised image captioning model with a visual concept detector which is trained on Visual Genome dataset \cite{krishna2016visualgenome}. Although they do not need paired image-caption corpus, a large-scale dataset with images and grounded objects annotations is also difficult to collect in any language.
Recently, cross-modal pivoted approaches are popularly used in solving zero-resource learning problems.
Chen \emph{et al.} \cite{chen2019wt, chen2019mt} propose to utilize images as pivots for zero-resource machine translation.
While, Gu \emph{et al.} \cite{gu2018pivot} and Lan \emph{et al.} \cite{lan2017fluencyguided} utilize language as pivot for cross-lingual image captioning.
Gu \emph{et al.} \cite{gu2018pivot} propose to train the English image captioning model on images paired with Chinese captions and English-Chinese parallel translation pairs. The model is performed in two steps through language pivoting, which has an inherent deficiency due to translation error accumulation. 
Lan \emph{et al.} \cite{lan2017fluencyguided} instead directly take advantage of the state-of-the-art translator to generate pseudo image-target caption pairs to train the captioning model. They propose to re-weight translated captions by language fluency to alleviate the disfluency errors brought about by the translator. However, in addition to disfluent sentences, the translation errors also contain fluent but visually irrelevant sentences, which are ignored in their works.

\begin{figure*}[t]
  \centering
  \includegraphics[scale=0.48]{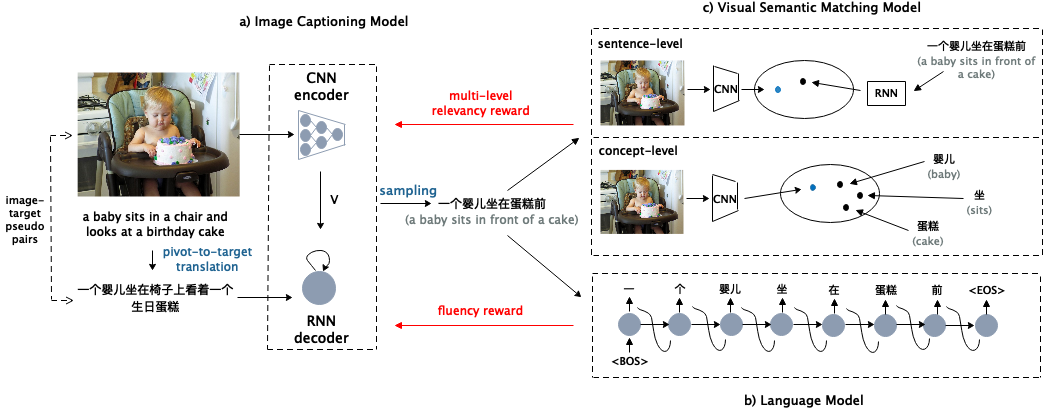}
  \vspace{-6pt}
  \caption{Illustration of the proposed SSR model framework, which consists of three components: a) the image captioning model trained on pseudo image-caption pairs; b) the language model to provide self-supervised fluency reward for the captioning model; c) the visual semantic matching model to provide self-supervised multi-level relevancy rewards. We add \textcolor{gray}{the English translation} below the sampled Chinese caption in brackets for better understanding.}
  \vspace{-6pt}
  \Description{Framework of our model.}
  \label{framework}
\end{figure*}

\section{Unpaired Cross-lingual Image Captioning with Self-supervision}
In this section, we will describe our self-supervised rewarding (SSR) model for unpaired cross-lingual image captioning.
We first present the overview of the model framework in Section~\ref{sec:method_overview}, which is based on reinforcement learning with two types of rewards to address the error accumulation problem in language-pivoted approaches.
Then in Section~\ref{sec:method_fluency} and Section~\ref{sec:method_relevancy}, we describe the proposed self-supervised fluency and relevancy rewards in details.

\subsection{Overview}
\label{sec:method_overview}
The goal of unpaired cross-lingual image captioning is to generate a natural language sentence to describe the image content in the target language without image-target caption pairs for training. 
We tackle this problem via a pivot language with the supervision from the help of image-caption pairs in the pivot language and the pivot-to-target translation model.
We refer to the pivot-to-target translation model as $f_{P \rightarrow T}$, and the image caption dataset in pivot language as $D_P = \{(I^{(i)}, d_P^{(i)})\}_{i=1}^{N}$, where $I^{(i)}$ refers to an image instance, $d_P^{(i)}$ refers to its corresponding sentence description in the pivot language, and $N$ is the total number of such image-caption pairs.
Therefore, although we don't have manually annotated image-caption pairs in the target language, we can generate pseudo pairs $D_T = \{(I^{(i)}, d_T^{(i)})\}_{i=1}^{N}$ based on $f_{P \rightarrow T}$ and $D_P$ where $d_{T}^{(i)} = f_{P \rightarrow T}(d_P^{(i)})$ for training.

If the translation model $f_{P \rightarrow T}$ is perfect, the pseudo pair $(I^{(i)}, d_T^{(i)})$ can be used as groundtruth to train the target image captioning model in a supervised way. Thus the unpaired cross-lingual image captioning can be converted to a standard image captioning task.
In this work, we employ the vanilla image captioning model based on an encoder-decoder framework \cite{vinyals2015showtell}. 
The encoder is a deep CNN \cite{kim2014cnn} to encode the image $I$ to a fixed-dimensional feature vector $v$. 
The decoder is a RNN \cite{Sepp1997lstm} to generate descriptions word by word conditioned on $v$. 
The whole model is optimized by maximizing the probability of generating each ``groundtruth'' caption words. The generation loss function can be expressed as:
\begin{equation}
\label{eqn:caption_cross_entropy}
    \mathcal{L}_{cap} = - \sum_{i=1}^N \sum_{j=1}^n\log P(w_{T,j}^{(i)}|w_{T,0:j-1}^{(i)},v^{(i)};\theta_{cap}),
\end{equation}
where $d_T^{(i)} = \{w_{T,1}^{(i)}, \cdots, w_{T,n}^{(i)}\}$, n is the length of $d_T^{(i)}$, $w_{T,0}^{(i)}$ is the sentence beginning signal <BOS>, and $\theta_{cap}$ is the parameters of the image caption model.

However, in reality, $f_{P \rightarrow T}$ is not perfect and can produce different translation errors such as disfluent translations or visually irrelevant translations as shown in Figure~\ref{example}.
Such translation errors can greatly deteriorate the image captioning performance because the training supervision $\mathcal{L}_{cap}$ for the captioning model relies on the translated sentences.
Therefore, extra supervisions are needed to mitigate the negative effects from $f_{P \rightarrow T}$, and provide accurate guidance for the caption generator.
In this paper, we utilize reinforcement learning framework to improve the caption performance by providing various rewards.

In reinforcement learning framework, the caption generation can be seen as a sequence decision process. 
The decoder of the captioning model can be seen as an agent, and the generation of each word can be seen as an action taken by the agent in each step.
When action decisions are finished, rewards will be fed back to the agent to ``tell'' how good these actions are.
The objective of reinforcement learning is to maximize expected rewards in the end of decision.
In order to address the disfluency and visual irrelevancy translation errors, we propose a fluency reward function $r_{flc}(\cdot)$ in Section~\ref{sec:method_fluency} and multi-level visual relevancy reward functions $r_{srlv}(\cdot)$ and $r_{crlv}(\cdot)$ in Section~\ref{sec:method_relevancy} to ``tell'' the captioning model how to improve the generated captions at both coarse and fine-grained levels.
Specifically, we adopt the ``self-critical'' \cite{rennie2017sc} reinforcement learning algorithm to train our model.
Firstly, we carry out Monte-Carlo sampling to sample a sentence $s_s^{(i)}$ and evaluate its caption quality with the proposed reward functions.
Then we utilize the greedy search algorithm to generate a sentence $s_b^{(i)}$ to provide baseline reward for the stability of reinforcement training.

Therefore, the joint optimization loss function to train the image captioning model consists of three parts:
\begin{equation}
\label{eqn:jointoptim}
    \mathcal{L} = \alpha\mathcal{L}_{cap} + \beta\mathcal{L}_{flc} + \gamma\mathcal{L}_{rlv}
\end{equation}
where
$\mathcal{L}_{flc}$ and $\mathcal{L}_{rlv}$ are the reinforcement learning objectives in fluency and relevancy aspects respectively; $\alpha, \beta$ and $\gamma$ are hyper-parameters, which are chosen according to the scale of these loss values and caption performance on the validation set.
Figure~\ref{framework} illustrates the overall framework of our proposed model.

\subsection{Self-supervised Fluency Rewards}
\label{sec:method_fluency}
In order to improve the fluency quality of generated captions, we employ self-supervision from mono-lingual corpus in the target language $S_T = \{s_T^{(i)}\}_{i=1}^{N}$ to provide the fluency reward.
We pre-train a language model on the mono-lingual corpus to evaluate the sentence fluency quality.
We utilize the LSTM as our language model which is trained to maximize the probability of generating target sentence $s_T^{(i)}$. 
Its loss function is expressed as:
\begin{equation}
\label{eqn:language_model}
    \mathcal{L}_{lm} = - \sum_{i=1}^N \sum_{j=1}^n\log P(w_{s,j}^{(i)}|w_{s, 0:j-1}^{(i)};\theta_{lm}),
\end{equation}
where $s_T^{(i)} = \{w_{s,1}^{(i)}, \cdots, w_{s,n}^{(i)}\}$, n is the length of $s_T^{(i)}$, and $\theta_{lm}$ is the parameter of the language model.

For a sampled sentence $s^{(i)} = \{ w_1^{(i)}, \cdots, w_n^{(i)} \}$ where $n$ is the sentence length, we take the log probability of generating $s^{(i)}$ by the language model as its fluency reward as follows:
\begin{equation}
\label{fluency}
    r_{flc}(s^{(i)}) = \frac{1}{n}\sum_{j=1}^n \log P(w_j^{(i)}|w_{0:j-1}^{(i)};\theta_{lm}).
\end{equation}

So the self-critical reinforcement loss function for fluency rewarding is formulated as:
\begin{equation}
\label{eqn:fluency_loss}
    \mathcal{L}_{flc} = - \sum_{i=1}^N (r_{flc}(s_s^{(i)})-r_{flc}(s_b^{(i)})) \sum_{j=1}^{n} \log P(w_{j}^{(i)}|w_{0:j-1}^{(i)}, v^{(i)};\theta_{cap}).
\end{equation}

\subsection{Self-supervised Relevancy Rewards}
\label{sec:method_relevancy}
Through the supervision from fluency reward, the caption model is ``taught'' to generate fluent captions in the target language. 
However, it cannot guarantee the generated captions are relevant to the given image, especially when the guidance from $\mathcal{L}_{cap}$ is wrong due to the semantically inconsistent translation errors.
Therefore, extra relevancy reward is highly required to let the captioning model know what is relevant to the image content and what is not.

We propose to learn a visual semantic matching model to evaluate the relevancy of the generated captions to the image based on the pseudo image-target caption pairs $D_T = \{(I^{(i)}, d_T^{(i)})\}_{i=1}^{N}$.
Although such pairs are noisy which might contain disfluent and visually irrelevant translation errors, there are also many content similar images whose descriptions are correctly translated, which enables accurate visual semantic matching.
we call this relevancy reward computed by the visual semantic matching model as ``self-supervised'' reward as no annotated image-target caption pairs are used.

In order to further mitigate noises in the translated sentences, we propose a multi-level visual semantic matching model (ML-VSE) which includes the image-sentence matching at the coarse level and the image-concept matching at the fine-grained level.
We use nouns and verbs in the caption sentence as the concepts, which play important roles to deliver semantic information of the sentence.
The concepts in the pseudo pairs can be more accurate than sentences since concepts don't suffer from disfluency errors and are easy to translate.
We describe the sentence-level and concept-level relevancy rewards computed by the two visual semantic matching models in details below.

\noindent\textbf{Sentence-Level Relevancy Reward.}
We provide the sentence-level relevancy reward via image-sentence matching. The image is encoded by $E_i$ which consists of a pre-trained CNN and a fully connected embedding layer to generate the image embedding vector $v_I$.
The caption sentence is encoded by $E_c$ which is a bi-directional GRU to generate the caption embedding vector $v_c$.
In order to project $v_I$ and $v_c$ in a common embedding space, we utilize the contrastive ranking loss with hard negative mining \cite{Faghri2018vse++} for training:
\begin{equation}
\label{eqn:sent_contrastive_loss}
\begin{split}
    \mathcal{L}(I,c)  = \max_{c'} [\Delta + s(v_{I}, v_{c'}) - s(v_{I},v_{c})]_{+}  \\
    + \max_{I'} [\Delta + s(v_{I'}, v_{c}) - s(v_{I},v_{c})]_{+},
\end{split}
\end{equation}
where $\Delta$ severs as a margin hyper-parameter, $[x]_+ \equiv \max(x,0)$, ($I$, $c$) is a pseudo image-caption pair, $c'$ is the negative caption given image $I$, and $I'$ is the negative image given caption $c$ in the mini-batch. The $s(\cdot)$ means the similarity function between two embedded vectors, which is the cosine similarity in our experiments.

After training, the image-sentence matching model is able to give captions that are relevant to the image higher similarity scores than irrelevant ones.
Therefore, our sentence-level visual relevancy reward for the generated caption $s$ of image $I$ is:
\begin{equation}
    r_{srlv}(s) = s(E_i(I), E_c(s)).
\end{equation}

\noindent\textbf{Concept-Level Relevancy Reward.}
Similarly to the image-sentence matching model, we utilize $E_i$ to encode the image to vector $v_I$ and encode the concept into the semantic vector $v_w$ with concept embedding matrix $E_w$.
The similar contrastive ranking loss is adopted to the joint image concept embedding space:
\begin{equation}
\label{eqn:concept_contrastive_loss}
\begin{split}
    \mathcal{L}(I,w)  = \max_{w'} [\Delta + s(v_{I}, v_{w'}) - s(v_{I},v_{w})]_{+}  \\
    + \max_{I'} [\Delta + s(v_{I'}, v_{w}) - s(v_{I},v_{w})]_{+}.
\end{split}
\end{equation}
The trained image-concept matching model can be used to measure the relevancy of the visual concept and the image.
However, the learned similarity score can be greatly influenced by the frequency statistics of concepts.
The frequent concepts in pseudo pairs are more likely to obtain high scores than infrequent ones, which biases the captioning model towards frequent concepts.
Hence, we normalize the similarity score by the prior probability of the concept in the dataset, so that the concept-level visual relevancy reward is computed as:
\begin{equation}
    r_{crlv}(w) = \delta(w) (s(E_i(I), E_w(w)) - \lambda p(w))
\end{equation}
where $(I, w)$ is the image-concept pair extracted from the pseudo image-caption pairs, $p(w)$ is the prior probability of $w$ which is its occurrence frequency, $\delta(w)$ denotes whether the word $w$ is a visual concept, and $\lambda$ is a hyper-parameter. 

Therefore, our multi-level self-critical loss to improve the visual relevancy of generated captions is as follows:
\begin{equation}
\label{eqn:relevancy_loss}
\begin{split}
    \mathcal{L}_{rlv} = - \sum_{i=1}^N \sum_{j=1}^{n} (r_{srlv}(s_s^{(i)})-r_{srlv}(s_b^{(i)}) + r_{crlv}(w_j^{(i)})) \\
    \cdot \log P(w_{j}^{(i)}|w_{0:j-1}^{(i)}, v^{(i)};\theta_{cap}).
\end{split}
\end{equation}

The overall training process of the proposed self-supervised rewarding model is presented in Algorithm \ref{alg::framework}.
\begin{algorithm}[h] 
  \caption{Training algorithm of the proposed self-supervised rewarding model for unpaired cross-lingual image captioning.}  
  \label{alg::framework}  
  \begin{algorithmic}[1]  
    \Require
      pivot image caption dataset $D_P$;
      pivot-to-target machine translation model $f_{P \rightarrow T}$;
      target language sentence corpus $S_T$;
    \State Generate pseudo image-target caption pairs $D_T$ based on $D_P$ and $f_{P \rightarrow T}$;
    \State Pre-train the target language model $\theta_{lm}$  based on $S_T$ with Eq (\ref{eqn:language_model});
    \State Pre-train $E_i, E_c, E_w$ in ML-VSE model based on $D_T$ with Eq (\ref{eqn:sent_contrastive_loss}) and Eq (\ref{eqn:concept_contrastive_loss}) respectively;
    \State Initialize $\theta_{cap}$ based on $D_T$ with Eq (\ref{eqn:caption_cross_entropy});
    \Repeat  
      \State select mini-batch $(I^{(i)}, d_{T}^{(i)}) \in D_T$;
      \State generate $s_s^{(i)}$ for $I^{(i)}$ via Monte-Carlo sampling;
      \State generate $s_b^{(i)}$ for $I^{(i)}$ via greedy search;
      \State compute fluency self-critic loss for $s_s^{(i)}$ by Eq (\ref{eqn:fluency_loss});  
      \State compute relevancy self-critic loss for $s_s^{(i)}$ by Eq (\ref{eqn:relevancy_loss});
      \State update $\theta_{cap}$ with Eq (\ref{eqn:jointoptim});
    \Until{$\theta_{cap}$ converges}  
  \end{algorithmic}  
\end{algorithm}

\section{Experiments}
We evaluate the unpaired cross-lingual image captioning models in both English and Chinese languages.
For unpaired English image captioning, we utilize Chinese as pivot; while for unpaired Chinese image captioning, we utilize English as pivot.

\subsection{Evaluation Setting}
\label{sec:compared_methods}
\textbf{Datasets.}
We conduct experiments on the MSCOCO \cite{lin2014mscoco} and AIC-ICC \cite{Wu2017AIC} image caption datasets in this work.
The MSCOCO dataset is annotated in English, which consists of 123,287 images and 5 manually labeled English captions for each image.
We follow the public split \cite{lin2014mscoco} which utilizes 113,287 images for training, 5,000 images for validation and 5,000 images for testing.
The AIC-ICC (Image Chinese Captioning from AI Challenge) dataset contains 238,354 images and 5 manually annotated Chinese captions for each image.
There are 208,354 and 30,000 images in the official training and validation set in AI challenge.
Since annotations of the testing set are unavailable in the AIC-ICC dataset, we randomly sample 5,000 images from its validation set as our testing set.
We use ``Jieba'' \footnote{https://github.com/fxsjy/jieba} to tokenize Chinese captions. 
The words with frequency more than 4 are added to our vocabulary.
We truncate English captions longer than 20 and Chinese captions longer than 16.
The statistics of the two datasets are presented in Table \ref{statistic}. 

\begin{table}[h]
    \vspace{-6pt}
    \caption{Statistics of the datasets used in our experiments.}
    \vspace{-8pt}
    \begin{tabular}{c|c|c|c|c}
        \toprule
        \textbf{Dataset} & \textbf{Lang.} & \textbf{\# Images} & \textbf{\# Captions} & \textbf{\# Vocabulary} \\
        \midrule
        AIC-ICC & zh & 240K & 1200K & 7,654 \\
        MSCOCO & en & 123K & 615K & 10,368 \\
        \bottomrule
    \end{tabular}
    \label{statistic}
\end{table}

For unpaired English image captioning, the task is to generate captions in English for images from MSCOCO dataset while no English image-caption pairs are used. In this setting, we use Chinese as the pivot language and utilize the AIC-ICC Chinese image caption dataset.
For unpaired Chinese image captioning, the task is to generate captions in Chinese for images from AIC-ICC dataset while no Chinese image-caption pairs are used. In this setting, we use English as the pivot language and utilize the MSCOCO English image caption dataset.

\noindent\textbf{Compared Methods.}
We compare our proposed model with the following four baseline models:
\begin{itemize}
    \item Baseline: The vanilla captioning model \cite{vinyals2015showtell} trained on pseudo pairs $D_T$ with cross-entropy in Eq (\ref{eqn:caption_cross_entropy}) without any rewards.
    \item Baseline+: The vanilla captioning model trained on pseudo pairs $D_T$ with CIDEr as reward in the reinforcement learning framework \cite{rennie2017sc}.
    \item 2-Stage pivot Google model \cite{gu2018pivot}: It utilizes a two-stage pipeline for unpaired cross-lingual image captioning. The image-to-pivot captioning model is the vanilla caption model \cite{vinyals2015showtell} and the pivot-to-target MT model is the online Google translator.
    \item 2-Stage pivot joint model \cite{gu2018pivot}: It utilizes a two-stage pipeline, including a image-to-pivot captioning model and pivot-to-target MT model. The two models share the same word embedding and are jointly trained to alleviate translation errors on the image caption domain.
\end{itemize}

\noindent\textbf{Metrics.}
We utilize the standard caption evaluation metrics to assess the quality of caption sentences, including BLEU \cite{Papineni2002bleu}, METEOR \cite{Michael2014meteor} and CIDEr \cite{Vedantam2015cider}.
As an image tells a thousand words, above objective evaluation metrics may not be able to fully measure the caption quality from different aspects. We therefore carry out human evaluation to further assess the caption quality from the fluency and visual relevancy aspects.

\subsection{Implementation Details}
\textbf{Image Captioning Model.}
We extract activations from the last pooling layer of ResNet-101 \cite{He2016resnet} which is pre-trained on ImageNet as our image features.
We encode the image feature to a 512-dimensional vector to initialize the hidden state of LSTM decoder.
The LSTM decoder contains 1 layer with 512 hidden units.
The dimensionality of the word embedding is set as 512.
We use the special token <BOS> and <EOS> to represent the beginning and ending of sentences. 
At test time, a beam-search decoding with beam size of 10 is used to generate captions.
We use the state-of-the-art Baidu translation API \footnote{http://api.fanyi.baidu.com} as our translation model $f_{P \rightarrow T}$.

\noindent\textbf{Language Model for Fluency Rewards.}
For unpaired English image captioning, we use texts in the MSCOCO training set to train the English language model.
For unpaired Chinese image captioning, we use texts in the AIC-ICC training set to train the Chinese language model.
However, the mono-lingual corpus is not subject to these datasets.
We do an ablation study in Section \ref{sec:ablation_studies} to compare the performance of our SSR model with the language model trained on different corpus.
The language model is a one-layer LSTM with 512 hidden units.
After training, the language model is fixed to evaluate the fluency of target captions.

\noindent\textbf{ML-VSE for Relevancy Rewards.}
For the image-sentence matching model, we use ResNet-101 \cite{He2016resnet} pre-trained on ImageNet as the CNN image encoder and one-layer bi-directional GRU with 512 hidden units as the sentence encoder.
The dimensionality of image-sentence joint space is set to be 1024.
For the image-concept matching model, we extract nouns and verbs as visual concepts from the translated captions via stanford parsing tools \footnote{http://nlp.stanford.edu:8080/parser/index.jsp}. In total, there are 3,231 visual concepts for unpaired English image captioning and 9,107 visual concepts for unpaired Chinese image captioning. 
The dimensionality of image-concept joint space is set as 512.

\noindent\textbf{Training Details.}
We pre-train the image captioning model, language model and ML-VSE model using Adam optimizer \cite{kingma2015adam} with a batch size of 128. For the image captioning model, the initial learning rate is 4e-4, while for the language and ML-VSE model, the initial learning rate is 2e-4.
In the self-critical reinforcement training, we set the learning rate as 4e-5 and batch size of 256.
We set hyper-parameters $\alpha$, $\beta$, $\gamma$ and $\lambda$ to 0.05, 0.15, 1.0 and 0.5 respectively. 
A dropout of 0.3 is applied to all models during training to prevent over-fitting.

\begin{table*}
\centering
\caption{Performance comparison with baseline methods for unpaired English image captioning evaluated on the MSCOCO dataset and unpaired Chinese image captioning evaluated on the AIC-ICC dataset.}
\vspace{-8pt}
\label{tab:sota_comparison}
\begin{tabular}{c|c|cccccc}
\toprule
\textbf{Task} & \textbf{Method} & \textbf{Bleu@1} & \textbf{Bleu@2} & \textbf{Bleu@3} & \textbf{Bleu@4} & \textbf{Meteor} & \textbf{CIDEr} \\ 
\midrule
\multirow{5}{*}{\begin{tabular}[c]{@{}c@{}}Unpaired English \\ Image Captioning\end{tabular}} & Baseline & 42.7 & 21.4 & 10.2 & 5.2 & 13.5 & 14.5 \\
 & Baseline+ & 44.0 & 22.0 & 10.5 & 5.3 & 13.0 & 14.6 \\
 & 2-Stage pivot Google model \cite{gu2018pivot} & 42.2 & 21.8 & 10.7 & 5.3 & \textbf{14.5} & 17.0 \\
 & 2-Stage pivot joint model \cite{gu2018pivot} & 46.2 & 24.0 & 11.2 & 5.4 & 13.2 & 17.7 \\ \cmidrule{2-8}
 & \textbf{Our SSR} & \textbf{52.0} & \textbf{30.0} & \textbf{17.9} & \textbf{11.1} & 14.2 & \textbf{28.2} \\
 \midrule
\multirow{3}{*}{\begin{tabular}[c]{@{}c@{}}Unpaired Chinese \\ Image Captioning\end{tabular}} & Baseline & 41.1 & 23.9 & 13.0 & 7.1 & 21.1 & 11.5 \\
 & Baseline+ & 41.6 & 24.4 & 13.3 & 7.3 & 21.1 & 11.6 \\ \cmidrule{2-8}
 & \textbf{Our SSR} & \textbf{46.0} & \textbf{30.9} & \textbf{19.3} & \textbf{12.3} & \textbf{22.8} & \textbf{18.3} \\ 
 \bottomrule
\end{tabular}
\end{table*}

\begin{table*}
\centering
\caption{The contribution of different rewards for unpaired cross-lingual image captioning on MSCOCO and AIC-ICC datasets.}
\vspace{-8pt}
\label{tab:ablation_rewards}
\begin{tabular}{c|c|cccccc}
\toprule
\textbf{Task} & \textbf{Rewards} & \textbf{Bleu@1} & \textbf{Bleu@2} & \textbf{Bleu@3} & \textbf{Bleu@4} & \textbf{Meteor} & \textbf{CIDEr} \\
\midrule
\multirow{4}{*}{\begin{tabular}[c]{@{}c@{}}Unpaired English \\ Image Captioning\end{tabular}} & No Reward & 42.7 & 21.4 & 10.2 & 5.2 & 13.5 & 14.5 \\
 & $r_{flc}$ & 45.9 & 23.4 & 11.4 & 5.8 & 13.4 & 16.1 \\
 & $r_{flc}+r_{srlv}$ & 50.6 & 28.7 & 17.1 & 10.6 & 13.8 & 26.7 \\
 & \textbf{$r_{flc}+r_{srlv}+r_{crlv}$} & \textbf{52.0} & \textbf{30.0} & \textbf{17.9} & \textbf{11.1} & \textbf{14.2} & \textbf{28.2} \\
 \midrule
\multirow{4}{*}{\begin{tabular}[c]{@{}c@{}}Unpaired Chinese \\ Image Captioning\end{tabular}} & No Reward & 41.1 & 23.9 & 13.0 & 7.1 & 21.1 & 11.5 \\
 & $r_{flc}$ & 45.8 & 30.3 & 18.6 & 11.6 & 22.5 & 18.0 \\
 & $r_{flc}+r_{srlv}$ & \textbf{46.1} & 30.7 & 19.1 & 12.1 & 22.6 & \textbf{18.5} \\
 & \textbf{$r_{flc}+r_{srlv}+r_{crlv}$} & 46.0 & \textbf{30.9} & \textbf{19.3} & \textbf{12.3} & \textbf{22.8} & 18.3 \\ 
 \bottomrule
\end{tabular}
\end{table*}

\subsection{Comparison with the State-of-the-arts}
\label{sec:quantitative_results}
Table~\ref{tab:sota_comparison} presents the unpaired cross-lingual image captioning performance in English and Chinese languages from our proposed approach and the compared baselines.
The proposed self-supervised rewarding (SSR) model achieves the best performance among all methods across different languages and evaluation metrics.
The ``Baseline'' method trained with imperfect pseudo pairs is inferior to all other methods.
It demonstrates that translation errors in pseudo pairs can significantly deteriorate the captioning performance even if we have utilized the state-of-the-art translation model.
In the ``Baseline+'' method, although self-critical reinforcement learning algorithm is employed to train the model, the improvements over ``Baseline'' method is marginal since it directly utilizes the noisy translated captions to provide rewards.
Our model instead is enhanced with fluency rewards and both coarse- and fine-grained visual relevancy rewards in the reinforcement learning framework.
The comparison of our model with ``Baseline+'' proves that the contribution mainly comes from the proposed self-supervised rewards rather than the ``self-critical'' reinforcement training.

Our approach also outperforms the 2-stage models in Gu \emph{et al.} \cite{gu2018pivot}.
The 2-Stage pivot Google model takes the advantage of the state-of-the-art translation model but ignores the translation errors for unpaired image caption generation.
The 2-Stage pivot joint model addresses the translation domain mismatch by joint training but cannot generalize to using the state-of-the-art translation model.
To be noted, our model is also more efficient than the 2-stage models in the testing phase since we do not depend on the 2-stage pipeline for caption generation in the target language.

\subsection{Ablation Studies}
\label{sec:ablation_studies}
\noindent\textbf{Contributions of different rewards.}
In Table~\ref{tab:ablation_rewards}, we ablate the unpaired captioning performance on different self-supervised rewards.
The fluency reward alone improves the baseline method on both unpaired English and Chinese image captioning, which demonstrates that the proposed fluency reward can effectively improve the quality of generated caption sentences.
However, the fluency reward only promotes the fluency of sentence without considering the visual relevancy.
Combining the fluency reward with both sentence- and concept-level visual relevancy rewards achieves additional performance gains on both languages.
We notice that the improvements of visual relevancy rewards are larger on unpaired English image captioning than the unpaired Chinese image captioning.
Since the diversity of images in the Chinese pivot language AIC-ICC dataset is smaller than that in the MSCOCO dataset, the unpaired English image captioning trained on pseudo image-caption pairs on the AIC-ICC dataset is more likely to suffer from visual irrelevancy problems.
Therefore, our proposed visual relevancy rewards can benefit more for unpaired English image captioning.

\begin{table}
\centering
\caption{Cross-modal retrieval performance using the proposed sentence-level semantic matching model trained on the self-supervised pseudo English pairs on the AIC-ICC training set. R@k represents recall in top k for the cross-modal retrieval.}
\label{tab:sentence_level_vse}
\vspace{-8pt}
\begin{tabular}{c|cc|cc}
\toprule
 & \multicolumn{2}{c|}{\textbf{Image-to-Sentence}} & \multicolumn{2}{c}{\textbf{Sentence-to-Image}} \\
 & \textbf{R@1} & \textbf{R@10} & \textbf{R@1} & \textbf{R@10} \\ 
 \midrule
AIC-ICC val & 52.8 & 85.9 & 37.7 & 81.2 \\
MSCOCO test & 22.7 & 58.7 & 12.8 & 48.7 \\
\bottomrule
\end{tabular}
\end{table}

\begin{figure}
  \centering
  \includegraphics[scale=0.36]{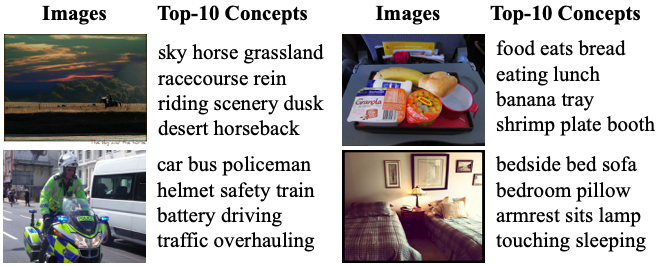}
  \vspace{-4pt}
  \caption{Top-10 predicted concepts for examples in MSCOCO test set.}
  \vspace{-8pt}
  \Description{Results of image-concept retrieval.}
  \label{retrieval}
\end{figure}

\noindent\textbf{Multi-level visual-semantic matching performance.}
We empirically evaluate the performance of ML-VSE model to demonstrate the reliability of the self-supervised relevancy rewards at the sentence- and concept-level.
We take the ML-VSE model trained for English captioning as an example.
We randomly select 1K images from the AIC-ICC validation set and MSCOCO testing set respectively to evaluate the performance of sentence-level semantic matching model, which is shown in Table~\ref{tab:sentence_level_vse}.
We notice that there exists a large performance gap between the MSCOCO testing set and AIC-ICC validation set, which can result from noises in pseudo pairs and image domain mismatch.
Therefore, additional fine-grained relevancy reward is requisite.
For the image-concept matching model, we visualize the top-10 predicted visual concepts for some images in the MSCOCO testing set in Figure \ref{retrieval}. 
As we can see, the predicted visual concepts are highly relevant to the image content, which cover diverse aspects such as object, action and scene.
Both results demonstrate the validity of our proposed relevancy guidance.

\begin{table} 
  \caption{English Image Captioning performance with language model trained on different mono-lingual corpus.}
  \vspace{-8pt}
  \begin{tabular}{c|c|c|c|c|c}
    \toprule
    \textbf{Corpus} & \textbf{\# Sents} & \textbf{B@3} & \textbf{B@4} & \textbf{Meteor} & \textbf{CIDEr} \\
    \midrule
    MSCOCO & 565K & 17.9 & 11.1 & 14.2 & 28.2 \\
    AIC-MT & 483K & 14.4 & 8.2 & 13.6 & 25.6 \\
    \bottomrule
 \end{tabular}
 \label{tab:corpus_comparision}
\end{table}

\begin{figure*}
  \centering
  \includegraphics[scale=0.45]{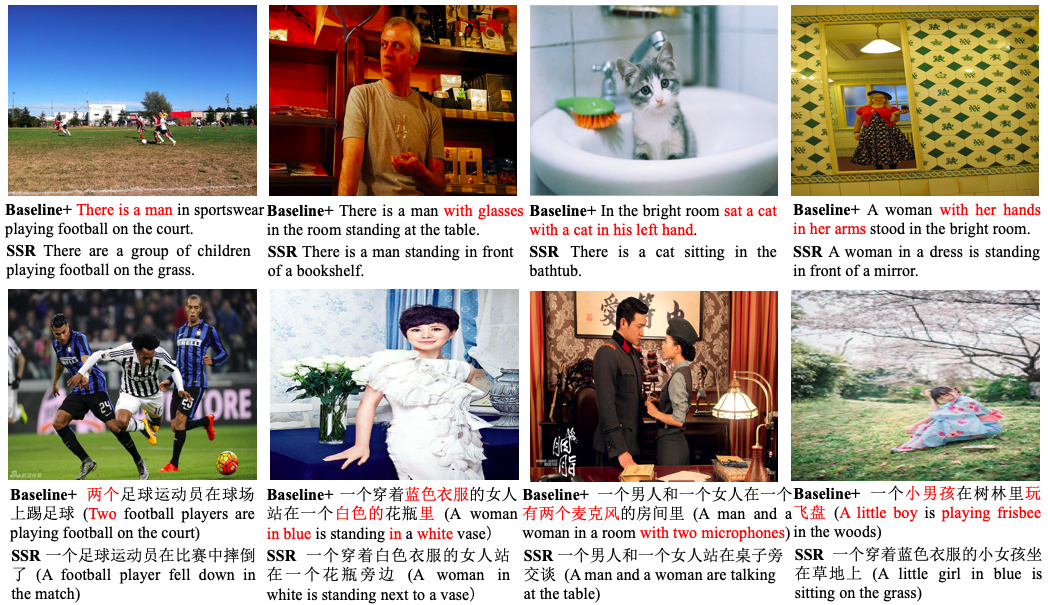}
  \vspace{-5pt}
  \caption{Examples of the English image captioning from the MSCOCO testing set, and Chinese image captioning from the AIC-ICC testing test. The errors in generated captions are marked in red.}
  \vspace{-10pt}
  \label{fig:visualization}
\end{figure*}

\noindent\textbf{Language model trained on different mono-lingual corpus.}
Although we utilize in-domain target corpus to train the language model in Table~\ref{tab:sota_comparison}, our SSR model can also benefit from other out-of-domain mono-lingual corpora which are easier to obtain in reality.
In Table~\ref{tab:corpus_comparision}, we present the unpaired English captioning performance with the language model trained on an out-of-domain corpus from AIC-MT \footnote{https://challenger.ai/competition/ect2018}.
Though using the out-of-domain mono-lingual corpus is not as effective as using in-domain data, it still achieves significant improvements over baseline models in Table~\ref{tab:sota_comparison}, which demonstrates the generalization ability of the proposed model to exploit different mono-lingual target corpora.

\noindent\textbf{Comparison with paired target image captioning.}
Table~\ref{supervisedresults} compares our proposed model with supervised mono-lingual image captioning models with different number of training pairs.
We can see that the number of paired image-caption data is critical for the supervised image captioning model.
Without sufficient pairs, the captioning performance drops significantly.
Our model, however, relies on no supervised image-caption pairs, but achieves performance comparable to the supervised mono-lingual captioning model with 4,000 pairs.

\begin{table}[h]
  \caption{Comparison between unpaired English image captioning and supervised English image captioning with different number of training pairs from MSCOCO dataset.}
  \vspace{-3pt}
  \begin{tabular}{c|c|c|c|c|c}
    \toprule
    \textbf{Approach} & \textbf{\# Imgs} & \textbf{\# Caps} & \textbf{B@4} & \textbf{Meteor} & \textbf{CIDEr} \\
    \midrule
    Baseline \cite{vinyals2015showtell} & 82,783 & 414,113 & 27.7 & 23.3 & 83.9 \\
    & 40,000 & 40,000 & 24.2 & 21.8 & 71.0 \\
    & 10,000 & 10,000 & 20.6 & 18.8 & 54.6 \\
    & 4,000 & 4,000 & 14.0 & 14.2 & 28.5 \\
    & 3,000 & 3,000 & 10.7 & 12.6 & 19.1 \\
    \midrule
    Our SSR & \textbf{0} & \textbf{0} & \textbf{11.1} & \textbf{14.2} & \textbf{28.2} \\
    \bottomrule
 \end{tabular}
 \label{supervisedresults}
\end{table}

\subsection{Human Evaluation and Qualitative Results}
Besides the quantitative evaluations in section \ref{sec:quantitative_results}, we also conduct human evaluation to verify the effectiveness of the proposed SSR model.
We take the unpaired English image captioning as an example.
We randomly select 1,000 images from the MSCOCO testing set, and recruit 10 workers who have sufficient English skills to evaluate the quality of generated captions from the ``Baseline+'' model and our SSR model.
Particularly, we measure the caption quality in the fluency and relevancy aspects. 
The fluency levels consist of 1-very poor, 2-poor, 3-barely fluent, 4-fluent, and 5-human like, 
and the relevancy levels consist of 1-irrelevant, 2-basically irrelevant, 3-partial relevant, 4-relevant, and 5-completely relevant.
Results in Table~\ref{humaneval} demonstrate that our approach can generate more fluent and visually relevant captions than the baseline model with the guidance of self-supervised rewards.
The example visualization results in Figure~\ref{fig:visualization} for both English and Chinese image captioning also confirm this.

\begin{table}[h]
  \caption{Human evaluation results on the MSCOCO 1K test.}
  \vspace{-8pt}
  \begin{tabular}{c|c|c}
    \toprule
    \textbf{Measure} & Baseline+ model & Our SSR model \\
    \midrule
    Fluency & 4.1 & 4.8 \\
    \midrule
    Relevancy & 3.3 & 3.8 \\
    \bottomrule
 \end{tabular}
 \label{humaneval}
\end{table}

\vspace{-8pt}
\section{Conclusions}
In this paper, we propose a novel language-pivoted approach for unpaired cross-lingual image captioning.
Previous language-pivoted methods mainly suffer from translation errors brought about by the pivot-to-target translation model, such as disfluency and visually irrelevancy errors.
We propose to alleviate negative effects from such errors by providing fluency and visual relevancy rewards as guidance in the reinforcement learning framework.
We employ self-supervisions from mono-lingual sentence corpus and machine translated image-caption pairs to obtain the reward functions.
Extensive experiments with both objective and human evaluations on both unpaired English and Chinese image captioning tasks demonstrate the effectiveness of the proposed approach.

\begin{acks}
This work was supported by National Natural Science Foundation of China (No. 61772535), Beijing Natural Science Foundation (No. 4192028), and National Key Research and Development Plan (No. 2016YFB1001202).
\end{acks}

\bibliographystyle{ACM-Reference-Format}
\balance
\bibliography{references}
\end{document}